\documentclass[pre,twocolumn,showpacs,showkeys,superscriptaddress,preprintnumbers,floatfix]{revtex4-1}

\usepackage{float}
\usepackage{etex}
\usepackage{ifpdf}
\usepackage{hyperref}
\usepackage{dcolumn}
\usepackage{url}
\usepackage{amsmath}
\usepackage{amscd}
\usepackage{amsfonts}
\usepackage{amssymb}
\usepackage{amsthm}
\usepackage{xfrac}
\usepackage{braket}
\usepackage{bm}   \usepackage{bbm}
\usepackage{verbatim}
\usepackage{mathtools}
\usepackage{stmaryrd}
\usepackage{xcolor}
\usepackage{setspace}
\theoremstyle{plain}    
\theoremstyle{plain}    
\theoremstyle{plain}    
\theoremstyle{plain}    
\theoremstyle{plain}    
\theoremstyle{plain}    
\theoremstyle{plain}    
\theoremstyle{plain}    
\theoremstyle{plain}    
\theoremstyle{plain}    
\theoremstyle{plain}    
\theoremstyle{plain}    
\theoremstyle{plain}    
\theoremstyle{plain}    
\theoremstyle{plain}    
\theoremstyle{plain}    
\theoremstyle{plain}

\newcommand{\eM}     {\mbox{$\epsilon$-machine}}
\newcommand{\eMs}    {\mbox{$\epsilon$-machines}}

\newcommand{\EMs}    {\mbox{$\epsilon$-Machines}}

\newcommand{\MeasAlphabet}  {\mathcal{A}}

\newcommand{\MeasSymbol}   { {X} }
\newcommand{\meassymbol}   { {x} }

\newcommand{\CausalState}   { \mathcal{S} }

\newcommand{\causalstate}   { \sigma }
\newcommand{\CausalStateSet}    { \bm{\CausalState} }
\newcommand{\AlternateState}    { \mathcal{R} }

\newcommand{\AlternateStateSet} { \bm{\AlternateState} }

\newcommand{\hmu}       {h_\mu}
\newcommand{\EE}        {{\bf E}}

\newcommand{\ProcessAlphabet}   {\MeasAlphabet}

\newcommand{\forward}{+}
\newcommand{\reverse}{-}
\newcommand{\forwardreverse}{\pm} 

\newcommand{\FutureCausalState} { {\CausalState}^{\forward} }

\newcommand{\PastCausalState}   { {\CausalState}^{\reverse} }

\newcommand{\lastindex}[2]{
  \edef\tempa{0}
  \edef\tempb{#2}
  \ifx\tempa\tempb
\edef\tempc{#1}
  \else
\edef\tempa{0}
    \edef\tempb{#1}
    \ifx\tempa\tempb
      \edef\tempc{#2}
    \else
      \edef\tempc{#1+#2}
    \fi
  \fi
  \tempc
}

\newcommand{\CSjoint}[1][,]{
   \edef\tempa{:}
   \edef\tempb{#1}
   \ifx\tempa\tempb
\ensuremath{\FutureCausalState\!#1\PastCausalState}
   \else
\ensuremath{\FutureCausalState#1\PastCausalState}
   \fi
}

\newif\ifpm
\edef\tempa{\forwardreverse}
\edef\tempb{\pm}
\ifx\tempa\tempb
   \pmfalse
\else
   \pmtrue
\fi

\newcommand{\Pe}{P_\text{e}}
\newcommand{\Pemin}{P_\text{e}^\text{min}}
\newcommand{\Hb}{H_\text{b}}
\newcommand{\Hbinv}{\Hb^{-1}}

\colorlet {R_color}    {blue}
\colorlet {k_color}    {black!30!green}

\def\clap#1{\hbox to 0pt{\hss#1\hss}}

\begin{document}

\title{
Complexity-calibrated Benchmarks for Machine Learning\\
Reveal When Next-Generation Reservoir Computer Predictions\\
Succeed and Mislead
}

\author{Sarah E. Marzen}
\affiliation{W. M. Keck Science Department of Pitzer, Scripps, and Claremont McKenna College, Claremont, CA 91711}

\author{Paul M. Riechers}
\affiliation{School of Physical and Mathematical Sciences, 
Nanyang Technological University,
637371 Singapore, Singapore}

\author{James P. Crutchfield$^*$}
\affiliation{Complexity Sciences Center and Physics Department,
University of California at Davis, One Shields Avenue, Davis, CA 95616,
chaos@ucdavis.edu
}

\date{\today}
\bibliographystyle{unsrt}

\begin{abstract}
Recurrent neural networks are used to forecast time series in finance, climate, language, and from many other domains. Reservoir computers are a particularly easily trainable form of recurrent neural network. Recently, a ``next-generation'' reservoir computer was introduced in which the memory trace involves only a finite number of previous symbols. We explore the inherent limitations of finite-past memory traces in this intriguing proposal. A lower bound from Fano's inequality shows that, on highly non-Markovian processes generated by large probabilistic state machines, next-generation reservoir computers with reasonably long memory traces have an error probability that is at least $\sim 60\%$ higher than the minimal attainable error probability in predicting the next observation. More generally, it appears that popular recurrent neural networks fall far short of optimally predicting such complex processes. These results highlight the need for a new generation of optimized recurrent neural network architectures. Alongside this finding, we present concentration-of-measure results for randomly-generated but complex processes. One conclusion is that large probabilistic state machines---specifically, large $\epsilon$-machines---are key to generating challenging and structurally-unbiased stimuli for ground-truthing recurrent neural network architectures.
\end{abstract}



\maketitle

\setstretch{1.1}

\newcommand{\Abet}{\ProcessAlphabet}
\newcommand{\MS}{\MeasSymbol}
\newcommand{\ms}{\meassymbol}
\newcommand{\SSet}{\CausalStateSet}
\newcommand{\St}{\CausalState}
\newcommand{\st}{\causalstate}
\newcommand{\MxSt}{\AlternateState}
\newcommand{\MxSSet}{\AlternateStateSet}
\newcommand{\mxst}{\mu}
\newcommand{\mxstt}[1]{\mu_{#1}}
\newcommand{\StartMS}{\bra{\delta_\pi}}
\newcommand{\Ipred}{I_\text{pred}}
\newcommand{\ISI} { \xi }

\newcommand{\ECT}{\widehat{\EE}}
\newcommand{\CCT}{\widehat{C}_\mu}

\newcommand{\gen}{g}
\newcommand{\FeatAlphabet}{\mathcal{F}}

\vspace{0.2in}

\section{Introduction}

Success in many scientific fields centers on prediction. From the early history of celestial mechanics we know that predicting how planetary objects move stimulated the birth of physics. Today, predicting neuronal spiking drives advances in theoretical neuroscience. Outside the sciences, prediction is quite useful as well---predicting  stock prices fuels the finance industry and predicting English text fuels social media companies. Recent advances in prediction and generation are so impressive (e.g., GPT-3) that one is left with the impression that time series prediction is a nearly solved problem. As we will show using randomness- and correlation-calibrated data sources, this hopeful state of affairs could not be further from the truth.

Recurrent neural networks \cite{lipton2015critical}, of which reservoir computers are a prominent and somewhat recent example \cite{schrauwen2007overview}, have risen to become one of the major tools for prediction. From mathematics' rather prosaic perspective, recurrent neural networks are simply input-dependent dynamical systems. Since input signals to a learning system affect its behavior, over time it can build up a ``memory trace'' of the input history. This memory trace can then be used to predict future inputs.

There are broad guidelines for how to build recurrent neural networks \cite{lipton2015critical} and reservoir computers that are good predictors \cite{schrauwen2007overview}. For instance, a linearized analysis shows that one wants to be at the edge of instability \cite{hsu2023strange}. However, a theory of how these recurrent neural networks work optimally is lacking; though see Ref.~\cite{krishnamurthy2022theory}. Recently, a new architecture was introduced for prediction called a ``next-generation reservoir computer'', whose memory trace intriguingly only included the last few timesteps of the input, while demonstrating low prediction error with simultaneously small compute power \cite{gauthier2021next}.

The general impression from these and many additional reports is that these recurrent neural networks have conquered natural stimuli, including language \cite{zhang2021commentary}, video \cite{zhou2020deep}, and even climate data \cite{chen2018applications}. They have certainly maximized performance on toy tasks \cite{hochreiter1997long,jaeger2012long} that test long memory.  This noted, it is unknown how far they are from optimal performance on the tasks of most importance, such as prediction of language, video, and climate.  We need a calibration for how far away they are from nearly-perfect prediction. And this suggests developing a suite of complex processes for which we know the minimal achievable probability of error in prediction

In the service of this goal, the following adopts the perspective that calibration is needed to understand the limitations inherent in the architecture of the next-generation reservoir computers and to understand how well state-of-the-art recurrent neural networks (including next-generation reservoir computers) perform on tasks for which optimal prediction strategies are known. This calibration is provided by time series data generated by a special type of hidden Markov model specialized for prediction called \emph{\eMs}.  We find, surprisingly perhaps, that large random multi-state \eMs\ are an excellent source of complex prediction tasks with which to probe the performance limits of recurrent neural networks.

More to the point, benchmarking on these data demonstrates that reasonably-sized next-generation reservoir computers are inherently performance limited: they achieve no better than a $\sim 60\%$ increase in error probability above and beyond optimal for ``typical'' \eM\ tasks even with a reasonable amount of memory. A key aspect of the calibration is that the optimalities are derived analytically from the \eM\ data generators, providing an objective ground truth. This increase in error probability above and beyond the optimal increases to $10^5\%$ if interesting \cite{Bial00a, Bial01a} stimuli are used. Altogether, we find that state-of-the-art recurrent neural networks fail to perform well predicting the high-complexity time series generated by large \eMs. In this way, next-generation reservoir computers are fundamentally limited.  Perhaps more surprisingly, a more powerful recurrent neural network \cite{hochreiter1997long} also has an increase in error probability above and beyond the minimum of roughly $50\%$ for these new prediction benchmarks.

Section~\ref{sec:Background} reviews reservoir computers, recurrent neural networks, and \eMs. 
Section~\ref{sec:PredictionErrorBounds}
derives a lower bound on the average rate of prediction errors.
Section~\ref{sec:Results} describes a new set of complex prediction tasks and surveys the performance of a variety of recurrent neural networks on these tasks. Section~\ref{sec:Discussion} draws conclusions and proposes new calibration strategies for neural network architectures. Such objective diagnostics should 
enable significant
improvements in recurrent neural networks.

\section{Background}
\label{sec:Background}

Section~\ref{subsec:eMs} describes \eMs\ and Sec.~\ref{subsec:RNN} lays out the setup of the typical recurrent neural network (RNN) and reservoir computer (RC).

\subsection{Complex processes and \eMs\ }
\label{subsec:eMs}

Each stationary stochastic process is uniquely represented by a predictive model called an \emph{\eM}. This one-to-one association is particularly noteworthy as it gives explicit structure to the space of all such processes. One can either explore the space of stationary processes or, equivalently, the space of all \eMs. This is made all the more operational, since \eMs\ can be efficiently enumerated \cite{John10a}.

In information theory they are viewed as process \emph{generators} and described as minimal unifilar hidden Markov chains (HMC). In computation theory they are viewed as process \emph{recognizers} and described as minimal probabilistic deterministic automata (PDA)~\cite{Shal98a, pfau2010probabilistic}. Briefly, an \eM\ has hidden states $\sigma\in\mathcal{S}$, referred to as \emph{causal states}, and generates a process by emitting symbols $x\in\mathcal{A}$ over a sequence of state-to-state transitions. For purposes of neural-network comparison in the following, we explore binary-valued processes, so that $\mathcal{A}=\{0,1\}$. \EMs\ are unifilar or ``probabilistic deterministic'' models since each transition probability $p(\sigma'|x,\sigma)$ from state $\sigma$ to state $\sigma'$ given emitted symbol $x$ are singly supported. More simply, there is at most a single destination state. In computation theory this is a deterministic transition in the sense that the model reads in symbols which uniquely determine the successor state. That said, these models are probabilistic as process generators: given that one is in state $\sigma$, a number of symbols $x$ can be emitted, each with emission probability $p(x|\sigma)$. In this way, these models represent stochastic languages---a set of output strings each occurring with some probability.

\begin{figure}
\centering
\includegraphics[width=0.45\textwidth]{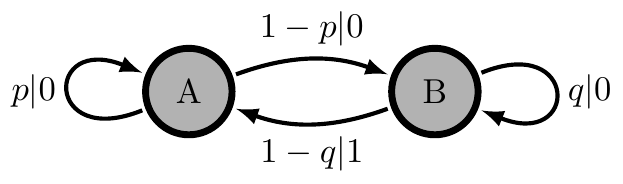}
\includegraphics[width=0.45\textwidth]{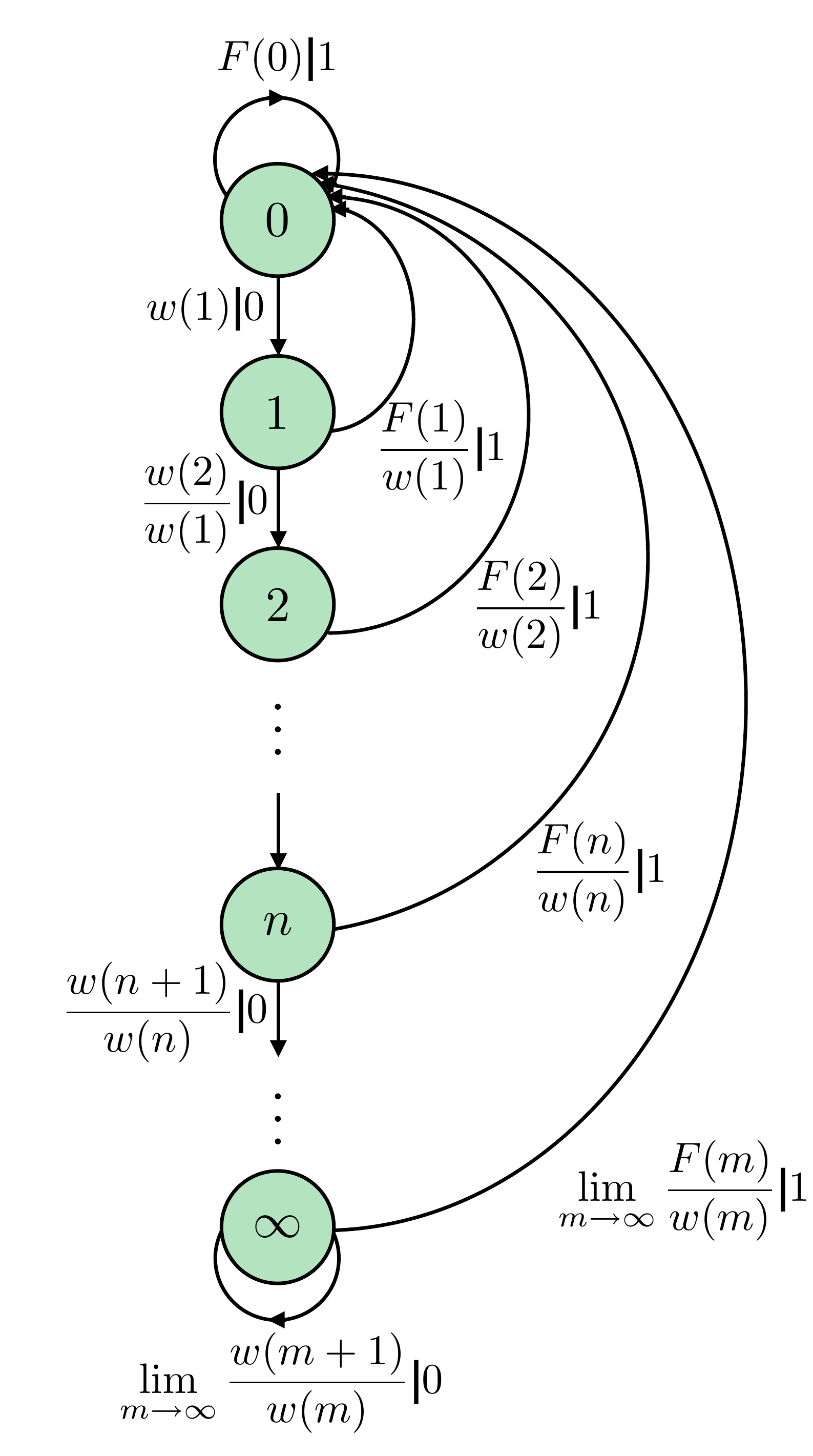}
\caption{At top, we see a nonunifilar hidden Markov model that is not an $\epsilon$-Machine because, when in state $A$, knowing that you have emitted a $0$ does not uniquely determine to which state one has transitioned.  At bottom, we see the corresponding $\epsilon$-Machine, for which in every state, knowing the emitted symbol uniquely determines the next state.  For this $\epsilon$-Machine, we have $F(n)=\begin{cases}
  (1-p)(1-q)(p^n-q^n) / (p-q) & p\neq q ~, \\
  (1-p)^2 n p^{n-1} & p=q ~.
  \end{cases}$ and $w(n)=\sum_{m=n}^{\infty} F(m)$ \cite{Marz14b}.  Note that both hidden Markov models generate an identical infinite-order Markov process: if one sees a past of all $0$'s, one has not ``synchronized'' to the internal hidden state of the $\epsilon$-Machine. Therefore, there is not a complete one-to-one correspondence between sequences of observed symbols and hidden states.}
\label{fig:0}
\end{figure}

While every stationary process has an \eM\ \emph{presentation}, it is usually not finite. An example is shown in Fig.~\ref{fig:0} \cite{Marz14b}. The finite HMC on the top is nonunifilar since starting in state $A$ and emitting a $0$ does not uniquely determine to which state one transits---either $A$ or $B$. The HMC on the bottom \emph{is} unifilar, since in every state, knowing the emitted symbol uniquely determines the next state.  Note that the \eM\ for the process generated by the finite nonunifilar HMC has an infinite number of causal states. Also, note that the process has infinite Markov order: if one sees a past of all $0$s, one has not ``synchronized'' to the \eM's internal hidden state \cite{Crut10a}. And, therefore, there is not a complete one-to-one correspondence between sequences of observed symbols and chains of hidden states. In contrast, with each step in the \eM\ presentation one inches closer to a one-to-one correspondence between observed symbols and hidden states---in reality, as close as possible.

In a way, a nonunifilar HMC is little more than a process generator \cite{lohr2009generative} for which the equivalent \eM\ presentation has an infinite number of causal states. In another sense, \eMs\ are a \emph{very} special type of HMC generator since the \eM's causal states actually represent clusters of pasts that have the same conditional probability distribution over futures \cite{Shal98a}. As a result, the casual states and so \eMs\ are \emph{predictive}.

Consider observing a process generated by a particular \eM\ and becoming synchronized so that you know the hidden state. (Now, this happens with probability $1$ but it does not always happen \cite{Crut10a}, as we just described with the nonunifilar HMC example.) Then you can build a prediction algorithm based on the known hidden state. The result, in fact, is \emph{the best possible prediction algorithm that one can build}. Moreover, the latter is simple: when synchronized to hidden state $\sigma$, you predict  the symbol $\arg\max_x p(x|\sigma)$.

This has one key consequence in our calibrating neural networks: the minimal attainable time-averaged probability $\Pemin$ of error in predicting the next symbol can be explicitly calculated as:
\begin{align}
\Pemin = \sum_{\sigma} \left[ 1 - \max_x p(x|\sigma) \right] p(\sigma)
  ~.
\end{align}
(The following considers binary alphabets, so that $ 1 - \max_x p(x|\sigma) = \min_x p(x|\sigma)$.)
We are also able to calculate the entropy rate $\hmu$ directly from the \eM\ \cite{Shal98a} via:
\begin{align}
\hmu & = H[X_0|\overleftarrow{X}_0] \nonumber \\
  & = - \sum_{\sigma} p(\sigma) \sum_x p(x|\sigma) \log p(x|\sigma)
  ~.
\end{align}
In contrast, until recently determining $\hmu$ for processes generated by nonunifilar HMCs was intractable. The key advance is that for these processes we recently solved Blackwell's integral equation \cite{blackwell1957entropy,Jurg20b}.

\subsection{Recurrent neural networks}
\label{subsec:RNN}

\begin{figure}
\centering
\includegraphics[width=0.45\textwidth]{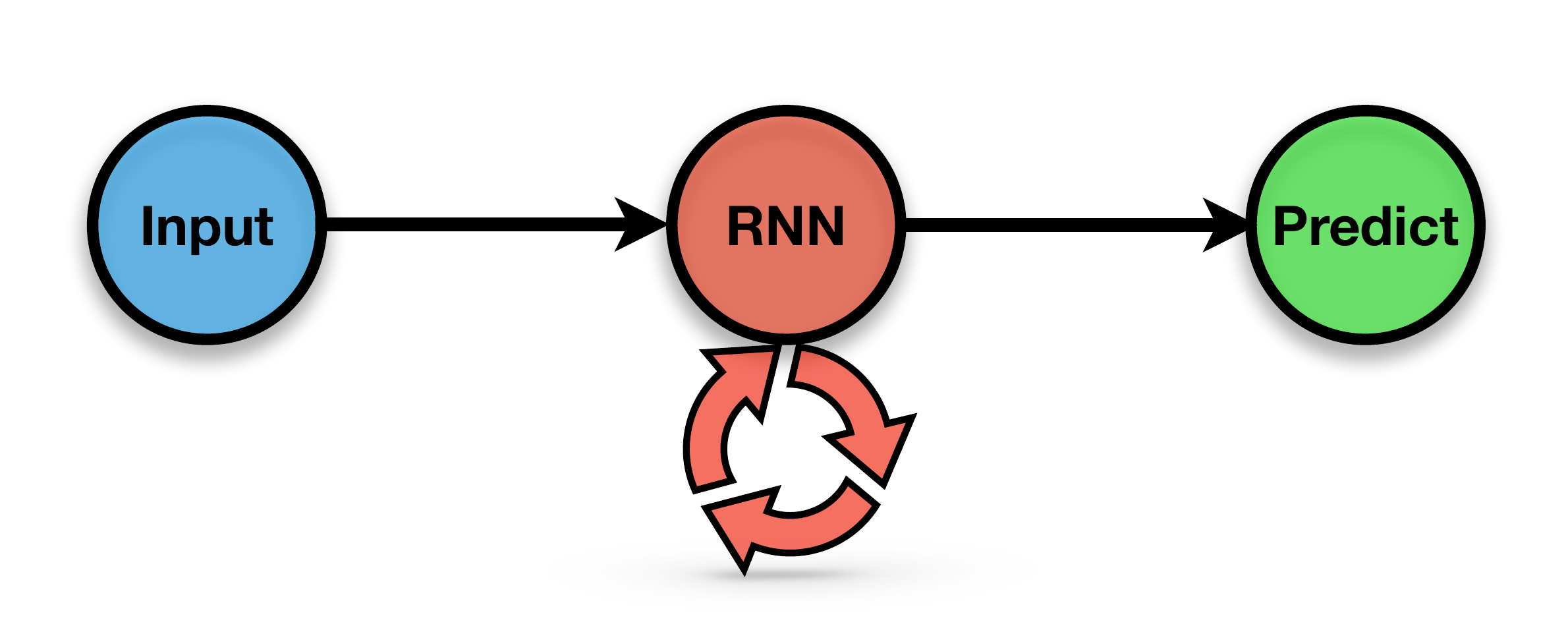}
\caption{A recurrent neural network (RNN) for which the future state of the recurrent node depends on its previous state and the current input. The present state of the recurrent node is then used to make a prediction.}
\label{fig:RNN}
\end{figure}

Let $s_t\in \mathbb{R}^d$ be the state of the learning system---perhaps a sensor---and let $x_t\in \mathbb{R}^N$ be a time-varying $N$-dimensional input, both at time $t$. Discrete-time recurrent neural networks (RNN) are input-driven dynamical systems of the form:
\begin{align}
s_{t+1} = f_{\theta}(s_t,x_t)
\label{eq:RNN}
\end{align}
where $f_{\theta}(\cdot ,\cdot )$ is a function of both sensor state $s_t$ and
input $x_t$ with parameters $\theta$; see Fig. \ref{fig:RNN}. These parameters are weights that govern how $s_t$ and $x_t$ affect future sensor states $s_{t+1}, s_{t+2}, \ldots$. Alternative RNN architectures result from different choices of $f$.  See below.  For simplicity, the following considers scalar time series: $N = 1$.

Generally, RNNs are hard to train, both in terms of required data sample size and compute resources (memory and time) \cite{pascanu2013difficulty}. RCs \cite{schrauwen2007overview, lukovsevivcius2009reservoir}, also known as \emph{echo state networks} \cite{jaeger2001short} and \emph{liquid state machines} \cite{maass2011liquid, maass2002real}, were introduced to address these challenges.

RCs involve two components. The first is a \emph{reservoir}---an input-dependent dynamical system with high dimensionality $d$ as in Eq.~\eqref{eq:RNN}. And, the second is a \emph{readout layer} $\hat{u}$---a simple function of the hidden reservoir state.  Here, the readout layer typically employs logistic regression:
\begin{align*}
P(\hat{u}|s) = \frac{e^{a_{\hat{u}}^{\top}s+b_{\hat{u}}}}{\sum_{\hat{u}'}e^{a_{\hat{u}'}^{\top}s+b_{\hat{u}'}}}
  ~,
\end{align*}
with regression parameters $a_{\hat{u}}$ and $b_{\hat{u}}$. To model binary-valued processes, our focus here, we have:
\begin{align*}
P(\hat{u}=1|s) = \frac{e^{a^{\top}s+b}}{1+e^{a^{\top}s+b}}
  .
\end{align*}
The regression parameters are easily trained and can include regularization if desired. Note that while $s$ was used as the input into the logistic regression probabilities, one can move to nonlinear readout by also using $ss^{\top}$ to inform the logistic regression probabilities.

The following compares several types of RNNs:
`typical' RCs,
`next generation' RCs,
and LSTMs.

\subsubsection{`Typical' RCs}

In this following, as a model of typical RCs,
a subset of RC nodes are updated linearly, while others are updated according to a $\tanh(\cdot)$ activation function.  Let $s = \begin{pmatrix} s^{\text{nl}} \\ s^{\text{l}} \end{pmatrix}$.  We have:
\begin{align*}
s^{\text{nl}}_{t+1} = \tanh \bigl( W^{\text{nl}} s^{\text{nl}}_t+v^{\text{nl}} x_t \bigr)
  ~,
\end{align*}
and
\begin{align*}
s^{\text{l}}_{t+1} = W^{\text{l}} s^{\text{l}}_t + v^{\text{l}} x_t ,
\end{align*}
where $v^\text{l/nl}$ controls how strongly the input affects the state and $W^\text{l/nl}$ are the weight matrices.
The weight matrices $W^\text{l/nl}$ are set to have a spectral radius $\sim 0.99$ to guarantee the RC fading-memory condition \cite{jaeger2001short}.

\subsubsection{`Next generation' RCs}

Next-generation RCs employ a simple reservoir that tracks some amount of input history and a more complex readout layer \cite{gauthier2021next} to improve accuracy over RC's universal approximation property. The reservoir records inputs from the last $m$ timesteps and, then, uses a readout layer consisting of polynomials of arbitrary order. Technically, next-generation RCs are a subset of general RCs in that a reservoir can be made into a shift register that records the last $m$ timesteps. 
As introduced in Ref.~\cite{gauthier2021next} next-generation RCs solve a regression task, but they can easily be modified to solve classification tasks. The following simply takes 
second-order
polynomial combinations of the last $m$ timesteps and uses those as features for the logistic regression layer. In other words, let $s_t = (x_t,x_{t-1},...,x_{t-m+1})$, a column vector, be the state of the reservoir; then we use $s_t$ and $s_t^{\top}s_t$ as input to the logistic regression.

\subsubsection{LSTMs}

In contrast, \emph{long short-term memory networks} (LSTM) \cite{hochreiter1997long} take a different approach by optimizing $f_{\theta}$ for training and for retaining memory. There, $s$ is  a combination of several hidden states and the update equations for the network are given in Ref. \cite{hochreiter1997long}.  An LSTM's essential components consist of linearly-updated \emph{memory cells} that make training easier and avoid exploding or vanishing gradients and a \emph{forget gate} that may improve performance by allowing the network to access a range of timescales \cite{krishnamurthy2022theory}.

\section{Prediction Error Bounds}
\label{sec:PredictionErrorBounds}

No matter the RNN, the conditional entropy of the next input symbol $X_t$ given the learning system's state $S_t$,
\begin{align*}
H[X_0|S_0] = - \sum_{s_t} p(s_t) \sum_{x_t} p(x_t|s_t) \log p(x_t|s_t) ~,
\end{align*}
places a fundamental upper bound on the RNN prediction performance through Fano's inequality:
\begin{align*}
H[X_0|S_0] \leq \Hb \left(\Pe \right) + \Pe \log \left(|\mathcal{A}|-1\right)
  ~.
\end{align*}
In this, $\Pe$ is the  time-averaged probability of making an error in \emph{predicting} the next symbol $x_t$ from RNN's  state $s_t$, and $\Hb$ is the binary entropy function. 
We have also invoked stationarity of the time series, to remove the dependence on $t$ in the steady-state operation of the RNN.
In particular, for a binary process where $|\Abet| = 2$:
\begin{align*}
\Pe \geq \Hbinv [ H(X_0 | S_0)] ~,
\end{align*}
where $\Hbinv$, defined on the domain $[0,1]$, is the inverse of $\Hb$ on its monotonically increasing domain $[0, 1/2]$.

In other words, the measure of RNN performance is given by a function of $H[X_0|S_0]$ that lower bounds $\Pe$, coupled with the minimal attainable probability of error calculable directly from the \eM\ as described in Sec.~\ref{sec:Background}.  The lower the model's conditional entropy, the better  prediction performance.  For any RNN, due to the Markov chain $S_0\rightarrow\overleftarrow{X}_0\rightarrow X_0$, this cannot be lower than $\hmu$---the entropy rate:
\begin{align*}
H[X_0|S_0] & \geq H[X_0|\overleftarrow{X}_0] \\
  & = \hmu
  ~.
\end{align*}
Notably, the next-generation RC takes into account only the last $m$ timesteps, so that:
\begin{align*}
H[X_0|S_0] & = H[X_0|\overleftarrow{X}_0^m] \\
  & = h_{\mu}(m),
\end{align*}
where the \emph{myopic entropy rate} $\hmu(m) \geq \hmu$ is discussed at length in Refs.~\cite{riechers2018spectral}.

\section{Results}
\label{sec:Results}

We are now ready to calibrate RNN and RC performance on the task of time-series prediction. First, we survey the performance of RCs when predicting a random sample of typical complex stochastic processes. Second, we explore RC performance on an ``interesting'' complex process---one from the family of memoryful renewal processes---hidden semi-Markov processes with infinite Markov order. Third and finally, we compare the prediction performance of RCs, next-generation RCs, and LSTM RNNs on a large suite of complex stochastic processes.

\subsection{Limits of Next-Generation RCs Predicting ``Typical'' Processes}
\label{sec:Results1}

We construct exemplars of ``typical'' complex processes by sampling the space of \eMs\ as follows:
\begin{itemize}
      \setlength{\topsep}{-2pt}
      \setlength{\itemsep}{-2pt}
      \setlength{\parsep}{-2pt}
\item An arbitrary large number of 
candidate
states is chosen for the HMC stochastic process generator. This parallels the fact that most processes have an infinite number of causal states \cite{pfau2010probabilistic,Marz17a};
\item For each ($\sigma$, $x$) pair, a labeled transition $\sigma \xrightarrow{ \, x \, } \sigma'$
is randomly generated, 
with the destination state $\sigma'$ chosen from the uniform distribution over candidate states;
\item Symbol emission probabilities $p(x|\sigma)$ are randomly generated from a Dirichlet distribution with uniform concentration parameter $\alpha$;
\item
We retain the largest recurrent component of this construction as our sample \eM.
\end{itemize}

Numerically, we find that 
approximately 20\% of the candidate states become transients in the constructed directed network, which are then trimmed from the final \eM.  
This number of transients strongly clusters around $20\%$ as the number of candidates grows large.  (Note that this is a topological feature, independent of $\alpha$.)
Moreover, this candidate network typically has a single recurrent component. 
Accordingly, the resulting causal states typically number about $80\%$ of the candidate states in our construction, as the number of candidate states grows large.

This results in a finite-state unifilar HMC or, equivalently, a presentation that can generate a process with a finite number of causal states. Interestingly, though, the process generated is \emph{usually} infinite-order Markov~\cite{Jame10a}. This can be seen from the  mixed-state presentation that describes optimal prediction \cite{riechers2018spectral,Jurg20b}, whose transient states of uncertainty generically maintain nonzero probability even after arbitrarily long observation time.~\footnote{This is typical even when the mixed-state presentation has a finite number of transient states. Adding a further challenge to the task of prediction, though, the mixed-state presentation typically has infinitely many transient states.}

\begin{figure}
\centering
\includegraphics[width=0.45\textwidth]{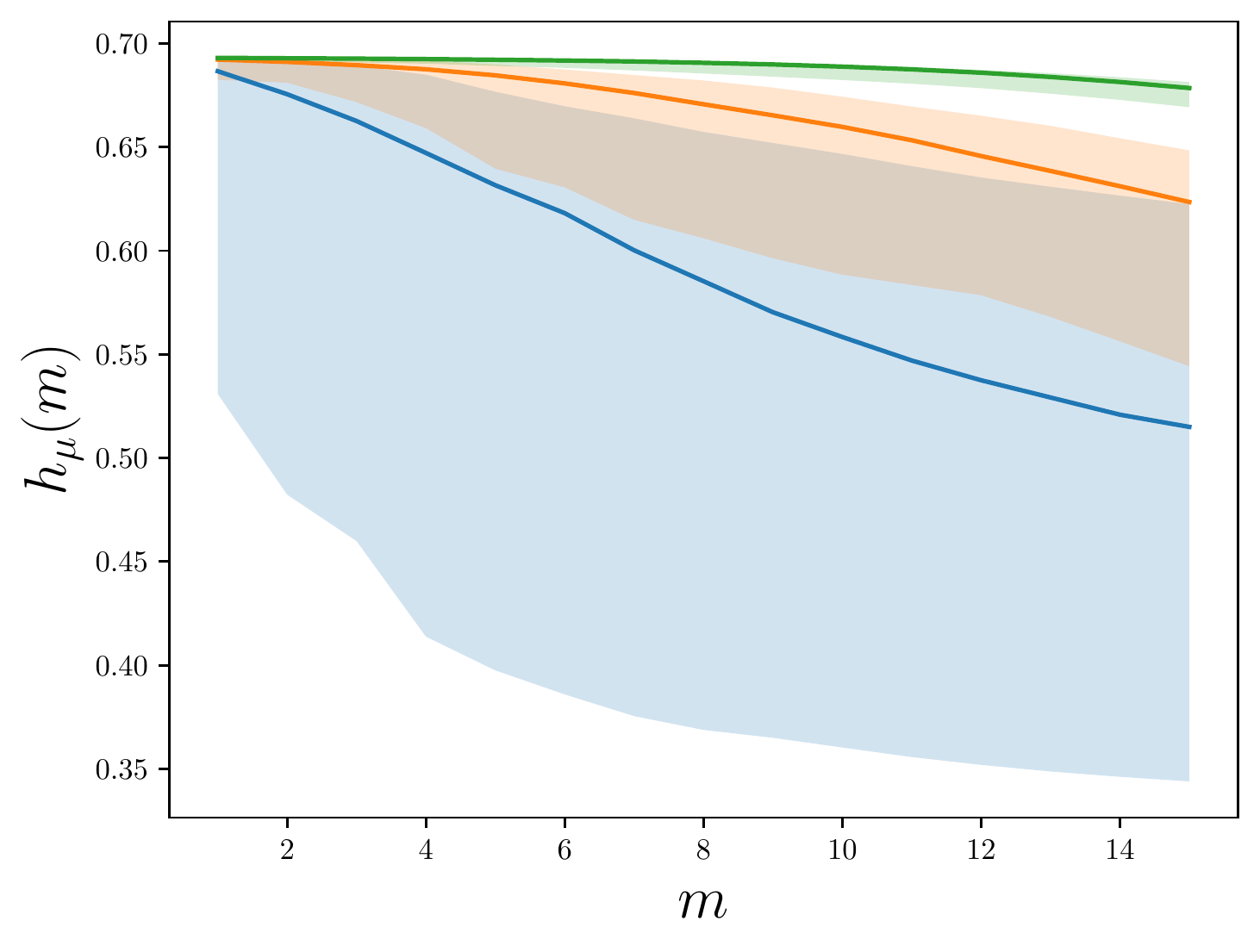}
\includegraphics[width=0.45\textwidth]{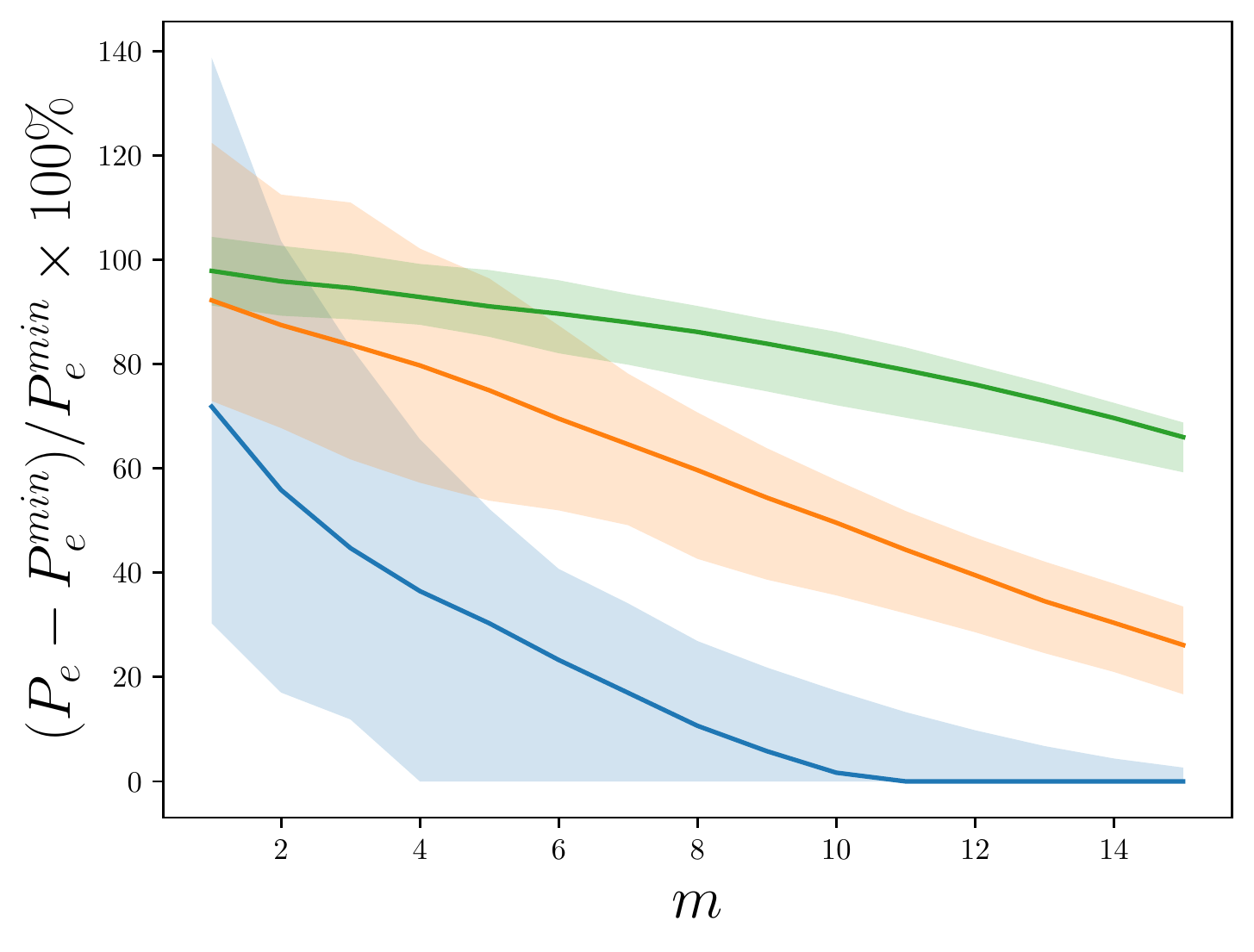}
\caption{(Top) Finite-length entropy rate $\hmu(m)$ in nats for typical random unifilar HMCs constructed with $30$ (blue), $300$ (orange), and $3000$ (green) candidate states as a function of the number of input timesteps $m$. (Bottom) Increase of the lower bound on the probability $\Pe$ of error from Fano's inequality, above and beyond $\Pemin$, with the same random 
unifilar HMCs as a function of the number of input timesteps $m$. Since occassionally the lower bound on this quantity fell below $0\%$, the maximum of $0\%$ and the quantity is used. $90\%$ confidence intervals are shown on both graphs.}
\label{fig:TypicalRCPerf}
\end{figure}

An expression for the myopic entropy rate $\hmu(m)$ was developed in Ref.~\cite{riechers2018spectral} that allows one to exactly compute $\hmu(m)$ from the generating \eM's mixed-state presentation. However, for binary-valued processes it was more straightforward to explicitly enumerate possible length-$m$ futures. Note, though, that this is impractical for the trajectory lengths used here if the emitted-symbol alphabet is larger than two. Figure~\ref{fig:TypicalRCPerf}(top) shows $\hmu(m)$ as a function of $m$, in the case that $\alpha = 1$. Figure~\ref{fig:TypicalRCPerf}(bottom) shows percentage increases in the $\Pe$ lower bounds for next-generation RCs above and beyond the minimal $\Pemin$, tracking prediction error lower bounds given by Fano's inequality in Sec.~\ref{sec:Results}.

Across this family of stochastic processes,
typical values of the myopic entropy rate $\hmu(m)$ and the entropy rate $\hmu$ exhibit a concentration of measure 
as the number of causal states grows large,
with values clustering around $1/2$ nat (not shown here). Typical values of the percentage increase in the $\Pe$ above and beyond the minimal $\Pemin$ show a concentration of measure, and the minimum probability $\Pemin$ of error cluster around $1/4$ (not shown here), reminiscent of the process-survey results reported by Ref.~\cite{Feld08a}.

A quick plausibility argument suggests that there is a genuine concentration of measure for these two quantities, using the formulae in Sec.~\ref{sec:Background}.  Roughly speaking, when the \eM\ generator has a large number of causal states, the transitions from any particular state have little effect on the stationary state distribution $p(\sigma)$. Hence, $\hmu$ and $\Pemin$ are roughly the sum of $N$ i.i.d.~ random variables. The Central Limit Theorem dictates for the concentration parameter $\alpha=1$ that $\hmu$ estimates should cluster around $1/2$ nat and that $\Pemin$ should cluster around $1/4$.
In contrast, $H[X_0]$ has the larger expected value of $\ln(2)$ nats, which becomes typical as the number of causal states grows large. The gradual decay of uncertainty from $\ln(2)$ to $1/2$ nat per symbol can only be achieved by predictors that (at least implicitly) synchronize to the latent state of the source via distinguishing 
long histories.

These typical processes are surprisingly non-Markovian, exhibiting infinite-range correlation. A process' degree of non-Markovianity is reflected in how long it takes for $\hmu(m)$ to converge to $\hmu$: how large must $m$ be to synchronize? Even after observing $m = 15$ symbols, these processes (with a finite but large number of causal states) are still $\sim 0.2$ nats away from synchronization. This convergence failure contributes to a minimal probability of error that cannot be circumvented \emph{no matter the cleverness in choosing the RC nonlinear readout function}.

\subsection{Limits of Next-Generation RCs Predicting an ``Interesting'' Process}
\label{sec:Results2}

References \cite{Bial00a, Bial01a} define complex and thus ``interesting'' processes as those that have infinite mutual information between past and future---the so-called ``predictive information'' or ``excess entropy''. The timescales of predictability are revealed through the growth $\Ipred(m)$ as longer length-$m$  blocks of history and future are taken into account. The \emph{predictive information} is:
\begin{align*}
\Ipred(m) = \sum_{l=0}^m \left[ \hmu (l) - \hmu \right]
  ~.
\end{align*}
And so, its growth rate is:
\begin{align*}
\Ipred(m+1) - \Ipred(m) & = \!\! \sum_{l=0}^{m+1} \!\! \left[\hmu(l)-\hmu\right]
  - \!\! \sum_{l=0}^m \!\! \left[ \hmu(l)-\hmu\right] \\
& = \hmu(m+1)-\hmu
  ~.
\end{align*}
That is:
\begin{align*}
\hmu(m+1) &= \hmu + \Ipred(m+1) - \Ipred(m)
  ~.
\end{align*}
The gap between $\hmu(m+1)$ and $\hmu$ quantifies the excess uncertainty in the next observable,  due to observation of only a finite-length past. This is governed by $\Ipred(m+1) - \Ipred(m)$ in discrete-time processes or, analogously, by $d \Ipred(t) / dt$ in continuous-time processes.

What constitutes an acceptable increase in prediction error above and beyond $h_{\mu}$? The intuition for this follows from inverting Fano's inequality to determine the additional conditional entropy implied by a substantial increase in the probability of error.

To illustrate this, we turn to an interesting process that has a very slow gain in predictive information---the discrete-time renewal process shown in Fig.~\ref{fig:0}(Bottom), with survival function:
\begin{align*}
w(n) = \begin{cases} 1 & n = 0 \\ n^{-\beta} & n\geq 1 \end{cases}
  ~.
\end{align*}
Discrete- and continuous-time renewal processes are encountered broadly---in the physical, chemical, biological, and social sciences and in engineering---as sequences of discrete events consisting of an event type and an event duration or magnitude. An example critical to infrastructure design occurs in the geophysics of crustal plate tectonics, where the event types are major earthquakes tagged with duration time, time between their occurrence, and an approximate or continuous \emph{Richter magnitude} \cite{Akim10a}. Another example is seen in the history of reversals of the earth's geomagnetic field \cite{Clar03a}. In physical chemistry they appear in single-molecule spectroscopy which reveals molecular dynamics as hops between conformational states that persist for randomly distributed durations \cite{Li13a,Li08a}. A familiar example from neuroscience is found in the spike trains generated by neurons that consist of spike-no-spike event types separated by \emph{interspike intervals} \cite{Marz14e}. Finally, a growing set of renewal processes appear in the quantitative social sciences, in which human communication events and their durations are monitored as signals of emergent coordination or competition \cite{Darm13a}.

At $\beta=1$, this discrete-time renewal process has  $\Ipred(m) \sim \log\log m$ \cite{Marz15a}. The minimal achievable lower bound is $\Pemin = 0.0001$. Due to an additional $\sim 0.1$ nats from not using an infinite-order memory trace and instead only using the last $m = 5$ symbols, the probability-of-error lower bound jumps to $0.02$. This is a percentage increase in probability of error of $10^4\%$ at $m = 11$ timesteps---about two and half orders of magnitude worse than that of a typical complex process. We emphasize these are fundamental bounds that no amount of cleverness can circumvent. While any nonlinear readout function might be chosen for a next-generation RC, the process' inherent complexity demands that an infinite-order memory trace be used for relatively good prediction.

\begin{figure}
\centering
\includegraphics[width=0.45\textwidth]{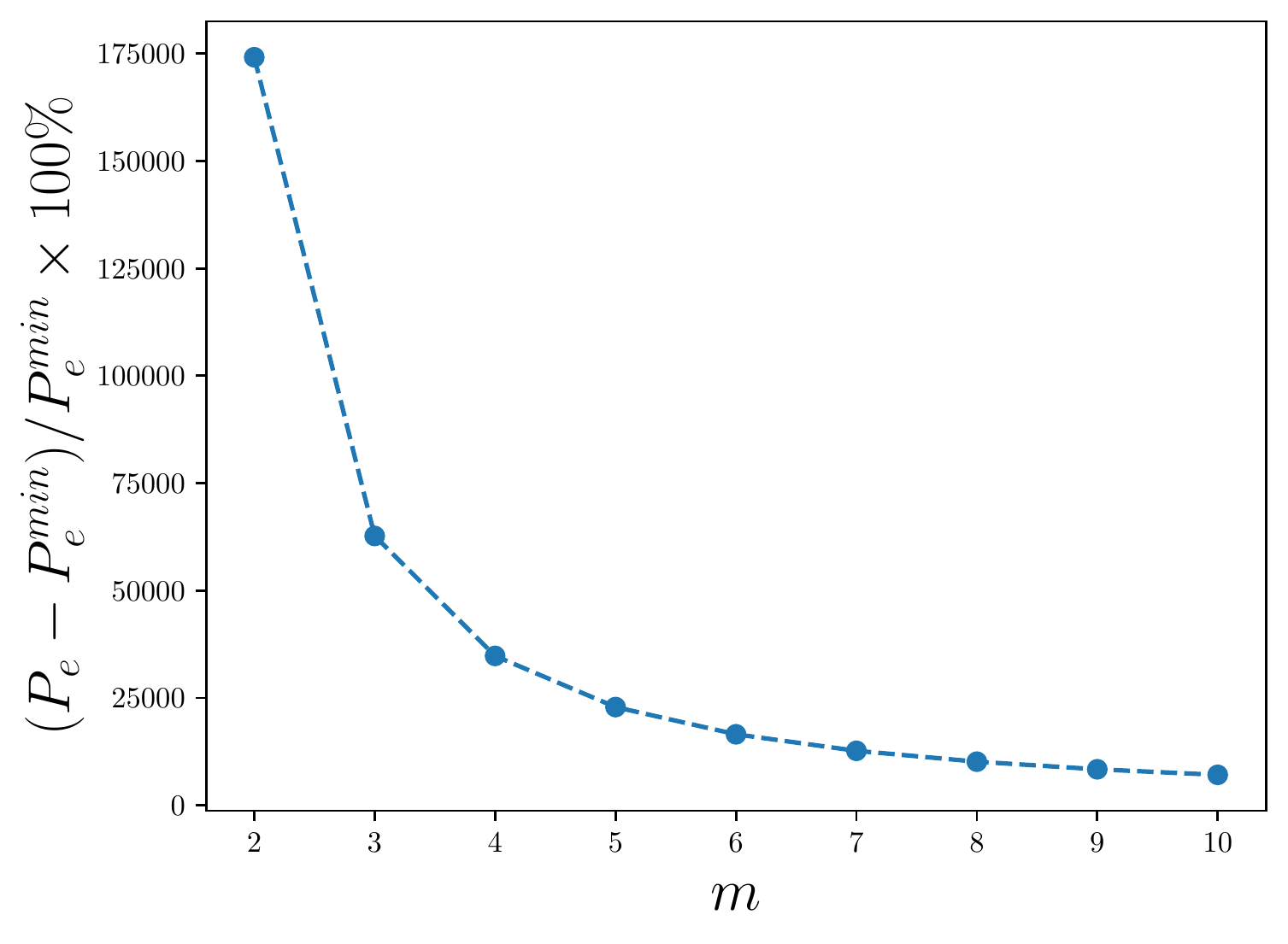}
\includegraphics[width=0.45\textwidth]{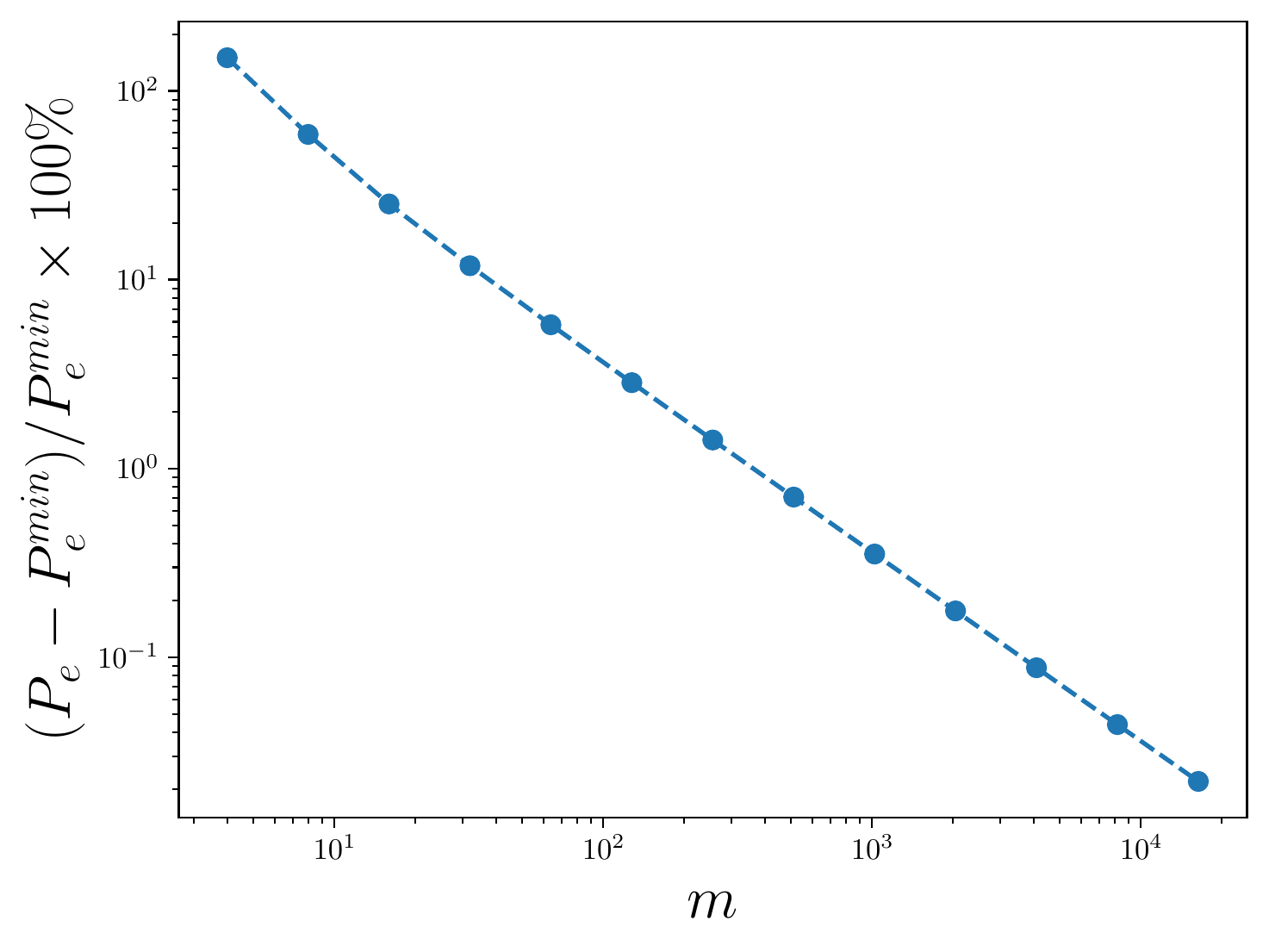}
\caption{Predicting  a discrete-time fractal renewal process with infinite excess entropy: (Top) Percentage increase in the lower bound for the probability of error $\Pe$ above and beyond the minimum using Fano's inequality as a function of time steps $m$.
(Bottom) Percentage increase in the lower bound for the probability of error $\Pe$ above and beyond the minimum using Fano's inequality as a function of time steps $m$ for a process such as that in Ref.~\cite{Marz15a}.}
\label{fig:2}
\end{figure}

References~\cite{Trav11b, Debo12a} constructed an HMC that ergodically \cite{Crut15a} generated $\Ipred(m) \sim \log m$.  For this process:
\begin{align*}
h_{\mu}(m+1) \approx h_{\mu} + \frac{1}{m}
  ~.
\end{align*}
Consider a process that has a ``typical'' entropy rate of $0.5$ nats, we can invert Fano's inequality---that is not necessarily tight---to find a lower bound on the probability of error with an infinite memory trace. Assuming this lower bound, the bound on the percentage increase of the probability of error above and beyond $\Pemin$ decays to $10\%$ only when the RC uses more than $m = 1000$ symbols.  See Fig. \ref{fig:2}(bottom).

\subsection{RCs, Next-Generation RCs, and State-of-the-Art RNNs Predicting Highly non-Markovian Processes}
\label{sec:Results3}

Knowing that there are fundamental limits to the next-generation RC's ability to predict processes forces the question: how well do next-generation RCs actually do at predicting these processes when using second-order polynomial readout? Moreover, do more traditional RCs and state-of-the-art RNNs do any better?

In all experiments, we are careful to hold the number of input nodes to the readout constant for a fair comparison.

\begin{figure}
\centering
\includegraphics[width=0.45\textwidth]{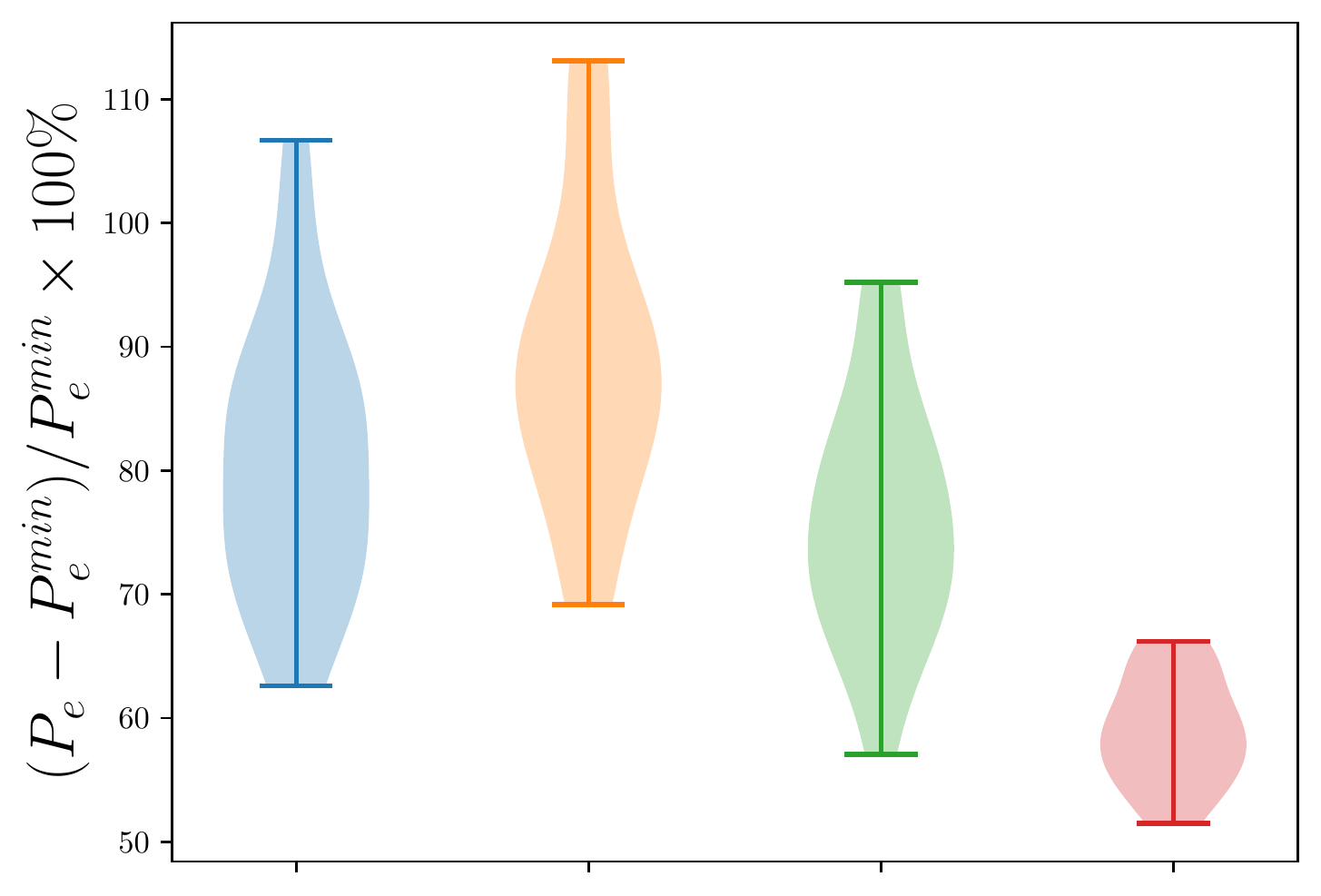}
\caption{Percentage increase in the probability of error of trained next generation RCs (green), trained RCs with linear readout (orange), trained RCs with nonlinear readout (blue), and trained LSTMs (red) above and beyond $\Pemin$ for $100$ \eMs\ with $300$ candidate states.  The next-generation RC has $10$ timesteps as input; the typical RC with nonlinear readout has $10$ nodes; the typical RC with linear readout has $110$ nodes; and the LSTM has $110$ nodes.  The number of nodes has been chosen so that the number of readout nodes is equivalent across machines. Note that these nearly saturate the lower bound provided by Fano's inequality.}
\label{fig:BigCompare}
\end{figure}

We now compare typical RCs with linear readout, typical RCs with nonlinear readout (second-order polynomial), and LSTMs to next-generation RCs on prediction tasks generated by the large \eMs\ of Sec.~\ref{sec:Results1}.  Although RCs with nonlinear readout and many more nodes outperform next-generation RCs, Fig.~\ref{fig:BigCompare} shows that when the number of readout nodes is held constant, next-generation RCs are indeed the best RC possible. This is expected from Ref.~\cite{gauthier2021next}. LSTMs beat all reservoir computers, however, as one can see from the red violin plot of Fig. \ref{fig:BigCompare} settling primarily on the lowest possible values of $(P_e-P_e^{min})/P_e^{min}\times 100\%$. This is somewhat expected since LSTMs optimize both the reservoir and readout, although the fact that they do is a testament to the fact that the successful training of the reservoir using backpropagation through time \cite{mozer1995focused}.

Figure \ref{fig:BigCompare}'s surprise is that all RNNs perform quite poorly, leaving at least $\sim 50\%$ increase in the probability of error above and beyond optimal, as one can see from the surprisingly large values on the $y$-axis, achieved at $m=10$ for the next-generation RC. This nearly saturates the lower bound on this percentage increase in the probability of error placed by Fano's inequality.

\section{Conclusion}
\label{sec:Discussion}

The striking advances made by RNNs in predicting a very wide range of systems---from language to climate---have not been accompanied by markedly improved explorations of how much structure they fail to predict. Here, we introduced and illustrated such a calibration.

We addressed the task of leveraging  past inputs to forecast future inputs, for any stochastic process. We showed that $\Pemin$---the minimal time-averaged probability of incorrectly guessing the next input, minimized over \emph{all} possible strategies that can operate on historical input---can be directly calculated from a data source's generating \eM. This provides a benchmark for all possible prediction algorithms. We compared this optimal predictive performance with a lower bound on various RNNs' $P_\text{e}$---the actual time-averaged probability of incorrectly guessing the next input, given the state of the model. We found that so-called next-generation RCs are fundamentally limited in their performance. And we showed that this cannot be improved on via clever readout nonlinearities.

In our comparison of various prediction models, we tested next-generation RCs with highly-correlated inputs that are challenging to predict. This input data was generated from large \eMs. The \eMs\ are the optimal prediction algorithm, and the minimal probability of error for these data are known in closed-form. Our extensive surveys showed, surprisingly, that models from RCs with linear readout to next-generation RCs of reasonable size to LSTMs all have a probability of prediction error that is $\sim 50\%$ greater than the theoretical minimal probability of error.

The fact that simple large random \eMs\ generate such challenging stimuli might be a surprise. Recently, though, it was reported that tractable \eMs\ can lead to ``interesting'' processes \cite{Bial00a, Bial01a}. We showed that these processes provide even more of a challenge for next-generation RCs.

Finally, next-generation RCs---that do indeed outperform typical RCs with the same number of readout nodes---are fundamentally limited in prediction performance by the nature of their limited memory traces.  We suggest that effort should be expended to optimize standard RCs that do not suffer from the same fundamental limitations---so that memory becomes properly incorporated and typical performance improves.

\vspace{1em}

\acknowledgments

The authors thank the Telluride Science Research Center for hospitality during
visits and the participants of the Information Engines Workshops there. JPC
acknowledges the kind hospitality of the Santa Fe Institute, Institute for
Advanced Study at the University of Amsterdam, and California Institute of
Technology for their hospitality during visits. This material is also based
upon work supported by, or in part by, the Air Force Office of Scientific
Research award FA9550-19-1-0411, Templeton World Charity Foundation grant
TWCF0570, Foundational Questions Institute and Fetzer Franklin Fund grant
FQXI-RFP-CPW-2007, U.S. Army Research Laboratory and the U.S. Army Research
Office grants W911NF-21-1-0048 and W911NF-18-1-0028, and U.S. Department of
Energy grant DE-SC0017324.

\section*{Data Availability}

The datasets used and/or analysed during the current study available from the corresponding author on reasonable request.

% \bibliography{chaos}

\end{document}